\documentclass{article}
\usepackage{ijcai22}

\usepackage{times}

\usepackage{soul}
\usepackage{url}
\usepackage[colorlinks=true,
            linkcolor=black,
            urlcolor=blue,
            citecolor=black]{hyperref}
\usepackage[utf8]{inputenc}
\usepackage[small]{caption}
\usepackage{graphicx}
\usepackage{amsmath}
\usepackage{booktabs}
\usepackage{xcolor}
\usepackage{amsthm}
\usepackage{mathtools}
\usepackage{amssymb}
\DeclarePairedDelimiter{\abs}{\lvert}{\rvert}
\usepackage{algorithm}
\usepackage{algpseudocode}
\definecolor{commentcolor}{RGB}{70, 150, 60}

\def\1#1{\mathcal{#1}}
\def\2#1{\mathbf{#1}}

\newcommand{\vpara}[1]{\vspace{0.2cm}\noindent\textbf{#1 }}

\title{Rethinking the Setting of Semi-supervised Learning on Graphs}

\author{
Ziang Li\thanks{indicates equal contribution.}\and
Ming Ding\footnotemark[1]\and
Weikai Li\and
Zihan Wang\and
Ziyu Zeng\and
Yukuo Cen\and
Jie Tang\\
\affiliations
Department of Computer Science and Technology, Tsinghua University\\
\emails
\{li-za19, dm18, liwk19, zhwang19, zengzy19, cyk20\}@mails.tsinghua.edu.cn
\emails
jietang@tsinghua.edu.cn
}

\begin{document}

\maketitle

\begin{abstract}
We argue that the present setting of semi-supervised learning on graphs may result in \textbf{unfair comparisons}, due to its potential risk of \textit{over-tuning hyper-parameters} for models.
In this paper, we highlight the significant influence of tuning hyper-parameters, which leverages the label information in the validation set to improve the performance. To explore the limit of over-tuning hyper-parameters, we propose ValidUtil, an approach to fully utilize the label information in the validation set through an extra group of hyper-parameters. With ValidUtil, even GCN can easily get high accuracy of 85.8\% on Cora.


To avoid over-tuning, we merge the training set and the validation set and construct an i.i.d. graph benchmark (IGB) consisting of 4 datasets. Each dataset contains 100 i.i.d. graphs sampled from a large graph to reduce the evaluation variance. Our experiments suggest that IGB is a more stable benchmark than previous datasets for semi-supervised learning on graphs. 
\footnote{Our code and data are released at \url{https://github.com/THUDM/IGB/}.
}
\end{abstract}

\section{Introduction}
Graph Neural Networks (GNNs)~\cite{gori2005new,kipf2016semi-GCN} have emerged as a heated field in machine learning in recent years. Among the widely known GNN models~\cite{kipf2016semi-GCN,Hamilton2017InductiveRL,Velickovic2018GraphAN,Grand,chen2020simple}, which one is the best? Most GNN papers demonstrate their performance on the task of semi-supervised learning on graphs following GCN, where a widely-used benchmark includes Cora, CiteSeer, and PubMed~\cite{Sen2008CollectiveCI}. 

This benchmark is competitive but unstable. For example, the accuracy of GCNII~\cite{chen2020simple} (SOTA method) on Cora is $85.5\%$ with a dropout rate of  $0.6$, but it will drop to $79.0\%$ if we slightly increase the dropout rate to $0.75$. In contrast, the reported accuracy of GCN is about 81.5\%.\footnote{The experiments can be  reproduced using the CogDL package~\cite{Cen2021CogDLAE}.} 
Previous researchers attributed this kind of instability to the \emph{small size} of the graphs and put forward larger benchmarks, e.g., OGB~\cite{Hu2020OpenGB} and HGB~\cite{hgb}. However, to the best of our knowledge, few works challenge the \emph{setting} of semi-supervised learning on graphs.   

A similar predicament of the unstable performance also exists in few-shot natural language understanding, where~\citeauthor{zheng2021fewnlu}~\shortcite{zheng2021fewnlu} recently found that models were over-fitting the validation set via hyper-parameters. Since the labels in the validation set are even more than those in the training set, the searched value of hyper-parameters becomes vitally important. 
Inspired by this finding, we hypothesize that the same reason could also, to some extent, account for the instability of semi-supervised learning on graphs, since the size of the validation set is also usually much larger than that of the training set (e.g., 140 training samples vs. 500 validation samples in Cora)~\cite{NELL}.
Meanwhile, recent GNNs show a trend of owning more hyper-parameters. While GCN has very few hyper-parameters,
GAT needs the accuracy on the validation set to determine its structures (e.g., the number of attention heads and the existence of a residual connection). PPNP~\cite{klicpera2019predict} further requires a global diffusion radius and a teleport probability $\alpha$. GDC~\cite{klicpera2019diffusion} searches a diffusion radius and a threshold for sparsification as hyper-parameters on the validation set, which is improved in ADC~\cite{zhao2021adaptive} by replacing the grid-search with gradient-based optimization for layerwise and channel-wise diffusion radii. This evolving path of GNNs suggests that GNN models utilize the validation set to a greater and greater extent and that the more explicit utilization improves and stabilizes the training. This benchmark leads to an unfair comparison favoring models with larger sizes of hyper-parameters.

The same phenomenon still exists in OGB~\cite{Hu2020OpenGB}, even though its percentage of validation set is much smaller than that of Cora. The participants~\cite{wang2021pairwise} find that directly merging validation set into the training set can significantly increase the performance, which is then only allowed on the collab dataset according to the updated OGB rules. Moreover, C\&S~\cite{huang2020combining} incorporates the labels in the validation set during label propagation and obtains a great improvement. All of them indicate that 
simply reducing the validation set ratio or to increase the graphs' size is not a satisfying solution to the problem. 

The findings urge us to rethink the setting of the semi-supervised learning on graphs and the meaning of validation set. 
On the one hand, the original motivation of introducing  validation set is to optimize the hyper-parameters, which cannot be directly optimized by usual methods, e.g., stochastic gradient descent (SGD). 
On the other hand, we have only two kinds of samples in real-world applications, labeled and unlabeled. 
We have to split out a part of the labeled data as the validation set to search for the best hyper-parameters.
This means it is a disadvantage instead of a merit to own a large set of hyper-parameters because they use a great quantity of labeled data as a validation set and result in a smaller training set providing real-world information. However, under the current setting, the extra and informative validation set encourages the model to equip itself with more hyper-parameters to fully utilize the labels in the validation set, which deviates from the real-world scenarios. In this work, we name the problem of ``using hyper-parameters to fit validation labels'' as \textbf{over-tuning}.

\textbf{Present work.} In this paper, we analyze the influence of the size of validation set and propose \textbf{ValidUtil}, a method to make any GNN fully utilize the label information in the validation set. \textbf{ValidUtil} explores the limit of over-tuning hyper-parameters. To avoid meaningless optimization towards the utilization of validation set and increase the stability of GNN benchmarks, we construct the i.i.d. graph benchmark (\textbf{IGB}) with the following two improvements:
\begin{itemize}
    \item \emph{Unify the training and validation set.} In IGB, the graphs have no pre-defined ratio of training and validation set but only labeled data and unlabeled data. Different models can freely split the labeled data into training and validation set based on the number of hyper-parameters. In this way, the over-tuning of hyper-parameters is discouraged. 
    \item \emph{Multiple i.i.d. graphs and diverse domains.} We have 4 datasets consisting of a co-authorship network, a social network, a knowledge graph, and a photo sharing network. In each dataset, we sample 100 subgraphs using a modified Random-Walk method. The sampled graphs are approximately i.i.d, while each gives a reliable evaluation of GNN performance. Therefore, we can obtain more stable metrics by averaging performances over them.
\end{itemize}

\section{The Risk of Over-tuning of Semi-supervised Learning on Graphs}
\subsection{Semi-Supervised Learning on Graphs}
\textbf{Definition. } Given an undirected graph $G=(V,E)$, where the node set $V$ contains $n$ nodes $\{v_1,...,v_n\}$ and $E$ is the edge set.
Each node $v_i$ is associated with a feature vector $\2x_i$ and a class label $y_i$. We denote the set of node labels as $\2Y$.
In the task of semi-supervised node classification on graphs (transductive), only a small part of node labels $\mathbf { Y } ^ { L } \subset \mathbf { Y }$ are given, and the rest label set $\mathbf { Y } ^ { U } = \mathbf { Y } - \mathbf { Y } ^ { L }$ needs to be predicted. Usually $ \abs{{ \mathbf{ Y } } ^ { L }} \ll \abs{\mathbf { Y } ^ { U }}$.

Here, we briefly introduce three widely used citation networks (i.e., Cora, CiteSeer, PubMed) for the analysis in this section. In these datasets, node features are bag-of-words representations of documents. Each dataset is a connected graph constructed based on the citation links between documents. Each dataset uses 20 training samples per class as labeled data in the semi-supervised setting. Table~\ref{tab:planetoid} shows the statistics of the three datasets.

\begin{table}[h]
\centering
\caption{Statistics of Cora / CiteSeer / PubMed datasets.}
\small
\label{tab:planetoid}
\setlength{\tabcolsep}{1.0mm}\begin{tabular}{l|rrccr}
	\toprule
	Dataset &  Nodes &  Edges & Split & Classes & Features \\
	\midrule
	Cora & 2,708 & 5,429 & 140 / 500 / 1,000 & 7&1,433  \\ 
	CiteSeer & 3,327 & 4,732 & 120 / 500 / 1,000 & 6 & 3,703 \\
	PubMed & 19,717 & 44,338 & 60 / 500 / 1,000 & 3 & 500 \\
	\bottomrule
\end{tabular}
\end{table}

\subsection{An Analysis of Over-tuning in Current GNNs}
In this section, we will investigate the over-tuning phenomenon in current GNNs. As discussed above, the hyper-parameters act as a tool to utilize the labels in the validation set. Therefore, the models' performance should improve as we increase the size of the validation set. We select five representative GNNs, GCN, GAT, APPNP, GDC-GCN, and ADC, and exhibit their accuracy on Cora with different sizes of validation set. The search scope of the hyper-parameters includes learning rate, hidden size, early stopping iteration, number of layers, the dropout rate, the diffusion radius of APPNP and GDC, the sparsification threshold of GDC, etc. We use grid search to find the best hyper-parameters for each model. Details about the search scopes are shown in the released codes.

\begin{figure}[h]
    \centering
    \includegraphics[width=0.9\linewidth]{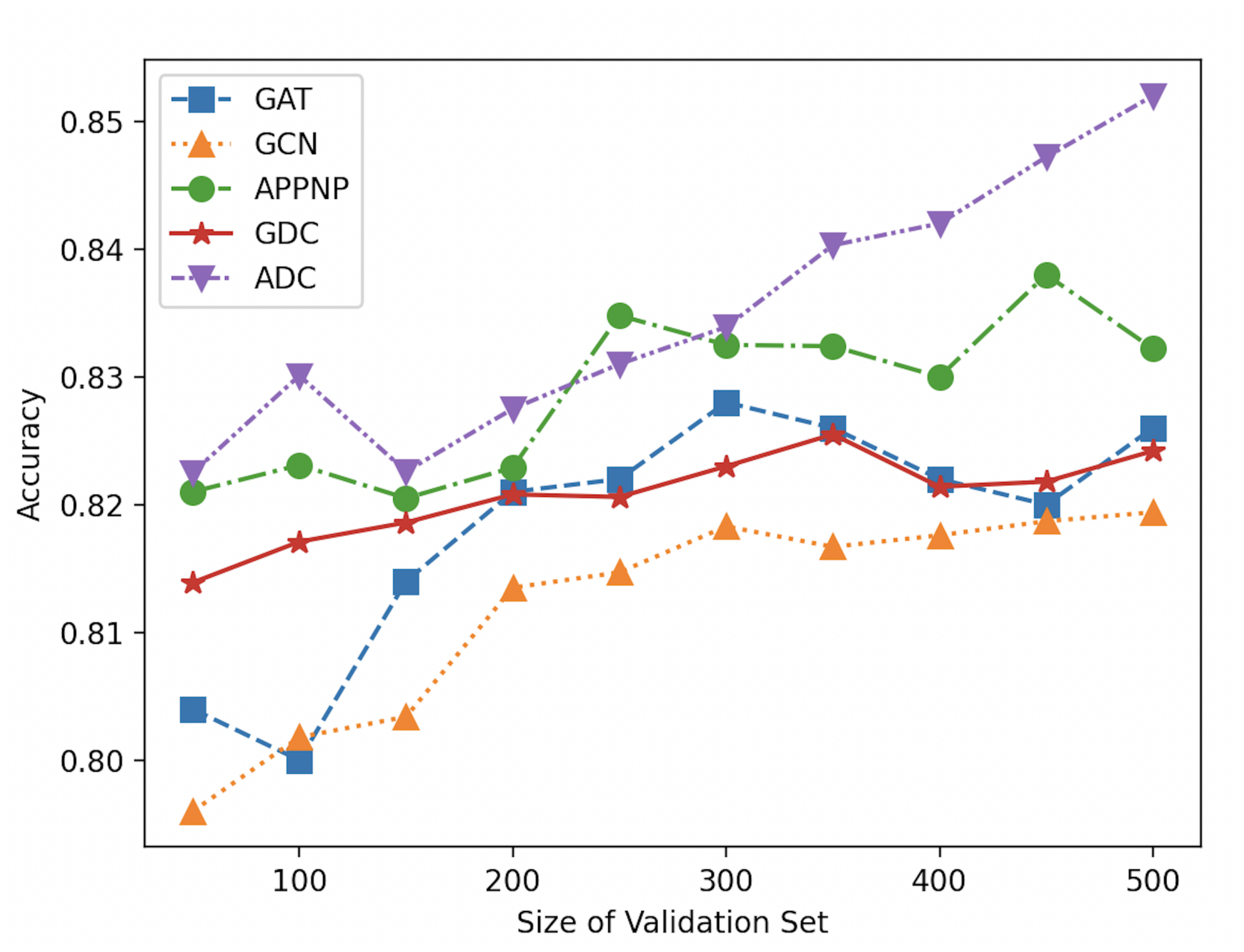}
    \caption{The accuracy of models with different size of the validation set on Cora. The test accuracy is the average of 20 runs with different random seeds.}
    \label{fig:varying_valid}
\end{figure}

We run the experiments on the public split~\cite{NELL} of Cora, ranging the size of validation set from 10 to 500 by hiding a part of the labels. For each validation size, we report the accuarcy on the test set after training the model with the best searched hyper-parameters. The results are demonstrated in Figure~\ref{fig:varying_valid}.

\begin{figure*}[t]
    \centering
    \includegraphics[width=\textwidth]{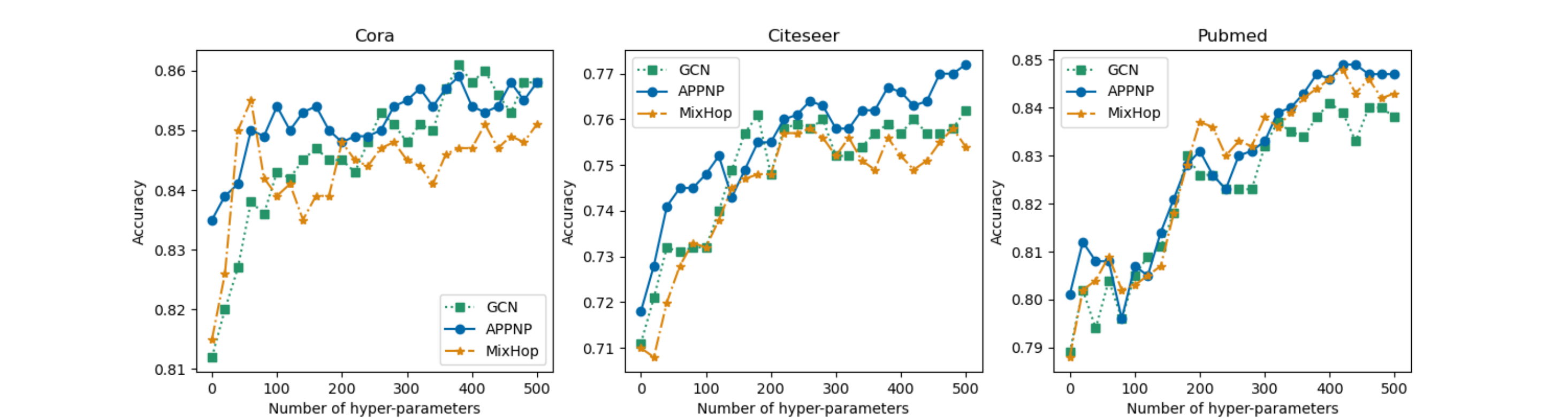}
    \caption{The test accuracy of ValidUtil plus GCN, GAT, MixHop and PPNP on Cora, Citeseer, and PubMed. The horizontal axis means the number of new hyper-parameters from ValidUtil, where 0 is equivalent to the original GNN without ValidUtil.}
    \label{fig:validutil}
\end{figure*}
Figure~\ref{fig:varying_valid} shows that the GNN models have a clear trend that the performance is usually better with a larger validation set. Since the validation set can only affect the model via the hyper-parameters, we can conclude that the model benefits from the validation labels with the help of hyper-parameters. The accuracy improvement is up to $1\%\sim3\%$ if we increase the size of validation set from 100 to 500, which is significant enough to suggest that the over-tuning already exists.

\subsection{ValidUtil: Exploring the Limits of Over-tuning}\label{sec:validutil}

Although the analysis above shows that over-tuning influences GNN models' performance, we wonder to what extent the influence could achieve. If the importance of validation labels is much smaller than that of model structures, the previous benchmarks will still be a proper choice for GNN benchmarks. If not, we should rethink and re-design the evaluation pipeline and datasets for semi-supervised learning on graphs. 

The most intuitive method to fully utilize the validation labels is to merge the validation set into the training set. However, this operation is universally acknowledged as invalid and is forbidden since it is a kind of data leakage. In this section, we put forward ValidUtil, a technique to mimic this operation via searching hyper-parameters. ValidUtil is \emph{not} really a method to improve GNNs, but more like a ``reduction to absurdity''. The full pipeline of ValidUtil is defined as follows and in Algorithm \ref{alg:valid_util}:


\begin{algorithm}[h]
    \caption{ValidUtil}
    \label{alg:valid_util}
    \textbf{Input} Graph $G$ with nodes in training/validation/test set. We denote their labels as $Y^{L}/Y_{valid}/Y_{test}$. There are $k$ classes. \\
    \textbf{Output} Accuracy of the GNN model on test set.
    \begin{algorithmic}[1]
    \State Train a GNN model $M$ with $Y^{L}$.
    \State Predict the labels of validation set with $M$ as $Y^P=\{y'_1, ..., y'_t\}$. 
    \State Add $t$ extra hyper-parameters $Y^T=\{\hat{y}_1, ..., \hat{y}_t\}$.
    \State Initialize $Y^T = Y^P$. 
    \For{$i$ from $1$ to $t$}
        \For{$l$ from $1$ to $k$}
            \State $\hat{y}_i = l$.            
            \State Train a GNN $M$ with $Y^{L} \cup Y^T$.
            \State $Acc_l$ = the accuracy of $M$ on validation set.
        \EndFor
        \State $l_{max}$  = $\underset{l \in \{1,...,k\}}{\arg\max} {Acc_l}$. 
        \If{$Acc_{l_{max}} > Acc_{l_{y_i'}}$}
        \State $\hat{y}_i$ = $l_{max}$
        \Else 
            \State $\hat{y}_i$ = $y'_i$
        \EndIf
    \EndFor
    
    \State Train a new GNN $M$ with $Y^{L} \cup Y^T$. 
    \State \Return the accuracy of $M$ on test set. 
    \end{algorithmic}

\end{algorithm}

\begin{enumerate}
    \item \textbf{Add hyper-parameters.} For any given model, we add $t$ extra hyper-parameters $\2Y^T=\{\hat{y}_1, ..., \hat{y}_t\}$, which refers to ``pseudo-labels of the $t$ nodes in the validation set''. These hyper-parameters affect the model training in a way that they act as a known label for the corresponding validation node; that is to say, the model will be trained on an augmented training set $\2Y^L \cup \2Y^T$. And since the labels of the validation set given to the model are not ground-truth labels, this is not a data leakage process.
    \item \textbf{Alternately optimize hyper-parameters.} The most common search method for hyper-parameters is grid search, but it is time-consuming when there are many hyper-parameters. Thus, we alternately search the best value of each $\hat{y}_i$ individually. In the beginning, each $\hat{y}_i$ is initialized as the predicted label $y'_i$ of the GNN without ValidUtil. We will search and fix the best $\hat{y}_i$ one by one, following the method described in the next paragraph. We set the dropout rate as 0 when searching for the best pseudo-labels. 
    \item \textbf{Search the best pseudo-label for each validation node.} To search for the best value of $\hat{y}_i$, we enumerate all possible values of $\hat{y}_i$ while keeping the other hyper-parameters unchanged. For each possible value, we train a GNN model using labels $\2Y^L \cup \2Y^T$ and select the best value according to the accuracy of the validation set. This is the standard process of searching a hyper-parameter.
    \item \textbf{Train the final model based on the best pseudo-labels.} After the pseudo-labels are determined, we can train on $\2Y^L \cup \2Y^T$, and use ordinary grid search to determine other hyper-parameters, including the dropout rate, and report the results on the test set. 
\end{enumerate}

\textbf{Analysis of Effectiveness.} The key reason for the effectiveness of ValidUtil is that in most cases, we can obtain the true label $y_i$ for validation node $v_i$ in step 3, which makes the final training equivalent to the training on the union of training and validation set. If the model is over-parameterized, it is powerful enough to overfit the predicted label of $v_i$ as the true label $y_i$ after sufficient training
\footnote{Some weak GNNs, e.g. vanilla GCN and GAT on CiteSeer, cannot well distinguish nodes in a strongly connected graph, and therefore cannot overfit the given labels. We add an additional self-loop for each node for GCN to solve the problem. This trick is already implemented in PyG~\cite{Fey/Lenssen/2019} by passing \texttt{improved=True} for GCNs. 
}.
Then in most cases, the highest accuracy is reached if and only if $\hat{y}_i = y_i$. We find that most GNNs models are powerful enough to overfit the pseudo-labels on Cora, Citeseer, and PubMed in practice. 
\begin{table}[h]
\small
\centering
\caption{Comparison between ValidUtil and the sota methods on Cora, Citeseer, and PubMed.}\label{tab:validutil_performance}
\begin{tabular}{lccc}
\toprule
     & Cora & Citeseer & PubMed \\
    \midrule
GCNII (sota Cora) & \textbf{85.5} & 73.4& 80.3\\
GRAND (sota Citeseer)& 85.4 & \textbf{75.4}& 82.7\\
SAIL (sota PubMed) & 84.6& 74.2 &\textbf{83.8}\\
\midrule
GCN + ValidUtil   &\textbf{85.8}& 76.0 & 83.8 \\
MixHop + ValidUtil  & 84.9 & 75.5 & 84.2\\
PPNP + ValidUtil  &\textbf{85.8}& \textbf{77.3}& \textbf{84.7}\\
\bottomrule
\end{tabular}
\end{table}

We demonstrate the performance of ValidUtil plus three GNN models, GCN, PPNP, and MixHop, in Figure~\ref{fig:validutil}. We find that even only $20\sim60$ hyper-parameters from ValidUtil can bring about a leap in performance for some models. When we add hyper-parameters for all the 500 nodes in the validation set, PPNP can achieve an accuracy much better than that of the sota methods in Table~\ref{tab:validutil_performance}.

\textbf{Remark.} Although ValidUtil works purely by utilizing the validation labels, it is totally valid under the current setting. If we treat the GNN+ValidUtil as a black-box model, the training process is quite normal. 
ValidUtil actually utilizes the labels with low efficiency, because each hyper-parameter can only learn the information of one node -- but this is enough to verify our hypothesis. \textbf{The current setting cannot prevent the validation labels from ``leaking'' during hyper-parameters tuning.} 
We believe that there exist some more efficient ways to define \emph{influential} hyper-parameters. These hyper-parameters could be entangled with the features or model structures, and they can acquire information from multiple validation labels. According to Figure~\ref{fig:varying_valid}, such influential hyper-parameters might already exist in some models and cannot be easily detected. Therefore, it is urgent to construct a new benchmark for semi-supervised learning on graphs to avoid over-tuning and fairly and robustly compare GNN models.

\section{IGB: An Independent and Identically Distributed Graph Benchmark}


\subsection{Overview}
Our new benchmark has two aims: avoiding over-tuning and being more robust.

To avoid over-tuning, we propose a new setting where there are only two sets of nodes, labeled and unlabeled. The model can use the labeled set in any way to train the best model and evaluate its performance on the unlabeled (test) set. If we need to search hyper-parameters, we can split out a part of the labeled nodes as the validation set. The over-tuning problem is eliminated because the validation labels are already exposed. This setting is closer to real-world scenarios and enables fair comparison between models with different sizes of hyper-parameters. To easily migrate the GNNs to this new setting, we will introduce a simple and powerful method to create validation set in section~\ref{sec:pipe}.

To construct a more robust benchmark, we expect models' performances to be stable for different random seeds. One of the most common methods in machine learning to reduce the variance of evaluation results is to test repeatedly and report the average performance. To achieve that, we expect to test a model's performance on multiple i.i.d graphs.   
However, how can we get multiple i.i.d. Cora-like graphs to evaluate the results? 

If we think about the construction of the citation networks such as Cora and Citeseer, we will find that the papers are crawled down by spiders from the Internet, which means these networks can be seen as sampled from the large real-world citation network. A similar assumption is already used in previous works~\cite{yang2020understanding} that the real-world graph data are sampled from a large underlying graph. To acquire i.i.d. graphs, we could sample ``again'' from an established graph. With appropriate sampling strategies, we can construct a group of i.i.d. graphs. The details about sampling are introduced in section~\ref{sec:sampling}.




\subsection{The Pipeline of Evaluation}\label{sec:pipe}
To solve the over-tuning problem, we have to update the pipeline of the task of semi-supervised learning on graphs. Following the real-world scenarios, we only split the nodes in a graph into two sets, labeled and unlabeled (with a ratio of 1:4 by default in IGB). The model can use the labeled set in any way to train the best model, and evaluate its performance on the unlabeled (test) set. A recommended method is as follows:
\begin{enumerate}
    \item Divide the labeled set into training and validation sets.\footnote{The best ratio may differ from model to model. In practice, we find that 1:1 is an appropriate ratio for most models.}
    \item Find the best hyper-parameters using grid search on the training and validation sets from the first step.
    \item Train the model with the best hyper-parameters on the full labeled nodes.
    \item Test the performance of the model from the third step on the unlabeled (test) sets.
    \item Repeat the above steps on each graph in a dataset and report the average accuracy.
\end{enumerate}
The first two steps aims to find the best hyper-parameters for the GNN model. We believe that this approach is suitable for many GNN models to get satisfying hyper-parameters. If there are other reasonable methods to decide the best hyper-parameters with the labeled set, they will also be encouraged to replace the first two steps in this pipeline. In this way, we can avoid over-tuning by directly exposing all the label information in the validation set later in the third step.

\subsection{Datasets}

IGB consists of four datasets: AMiner~\cite{AMiner}, Facebook~\cite{Facebook}, NELL~\cite{NELL}, and Flickr~\cite{Amazon}. Each dataset contains 100 undirected connected graphs, sampled from the original large graph according to the random walk method in section~\ref{sec:sampling}. 
We also report the average node \emph{overlap rate}, the ratio of common nodes to the total size of nodes for a pair of sampled graphs. 
The coverage rate is defined as the ratio of the union of the 100 sampled graph to the original large graph. 
Lower overlap rate and higher coverage rate are preferred. 
The statistics of the datasets are reported in Table \ref{tab:dataset_statistics}. 

\begin{table}[h]
    \centering
    \caption{Statistics of the datasets in IGB. }
    \label{tab:dataset_statistics}
    \scalebox{0.8}{
    \begin{tabular}{lcccc}
    \toprule 
        & AMiner& Facebook & NELL & Flickr \\  
    \midrule
     Average Nodes & 4,485 $\pm$ 26 & 3,475 $\pm$ 71 & 3,540$\pm$ 68 & 4,452 $\pm$ 31\\
     Edges & 5,000 & 5,000 & 5,000 & 5,000 \\
     Features & 3,883 & 128 & 10,000 & 500\\
     Classes & 8 & 4 & 164 & 7\\
     Original Size & 236,017 & 22,470 & 63,910 & 89,250 \\
     Overlap Rate & 0.083 & 0.339 & 0.146 & 0.119 \\
     Coverage Rate & 0.550 & 0.958 & 0.956 & 0.969 \\
    \bottomrule
    \end{tabular}
    }
\end{table}

\vpara{AMiner.}
The AMiner dataset is a co-authorship graph extracted from the AMiner system~\cite{AMiner}. Nodes represent authors, while edges mean co-authorship in at least one paper. Node features indicate the venues in which the author has publications. Specifically, each feature has 3,883 dimensions, and each dimension is 0 or 1, representing whether or not an author has had publications in the corresponding venue. Node labels represent the authors' main research fields.

\vpara{Facebook.}
The Facebook dataset is the Facebook Page-Page dataset from the paper~\cite{Facebook}. It is a graph of official Facebook pages. Nodes are official Facebook pages, while edges are mutual likes between the pages. Node features are extracted from the page descriptions. Node labels are one of the following 4 categories defined by Facebook: politician, governmental organization, television show, and company.

\vpara{NELL.}
The NELL dataset is a knowledge graph dataset generated from the NELL knowledge graph~\cite{NELL_raw}. Nodes represent entities, and edges represent relationships between two entities. Each node initially has a 61,278-dimension feature, a binary bag-of-words representation of entity descriptions. We only reserve the features of the most frequent 10,000 words for efficiency.

\vpara{Flickr.}
The Flickr dataset is a graph of photos uploaded to the Flickr website. Each node represents a photo, and an edge means two photos share some properties in common, such as from the same location or the same gallery. 500-dimension node features are bag-of-words representations of photos. The labels are one of 7 classes developed by \citeauthor{Amazon} from 81 original tags.





\subsection{Sampling Algorithm}\label{sec:sampling}
The simplest way to make the subgraph's node label distribution similar to that of the original graph is vertex sampling. However, it does not meet our expectations because it generates unconnected subgraphs. To obtain nearly i.i.d. subgraphs for our benchmark, we must carefully design the sampling strategy and principles. Specifically, we expect the sampling strategy to have the following properties:

\begin{enumerate}
    \item The sampled subgraph is a connected graph.
    \item The distribution of the subgraph's node labels is close to that of the original graph.
    \item The distribution of the subgraph's edge categories (edge category is defined by the combination of its two endpoints' labels) is close to that of the original graph.
\end{enumerate}

The first property can be well satisfied by the Random-Walk (RW) algorithm. When performing RW on an undirected graph $G = (V,E)$, we start sampling from node $u = n_0$, and the following nodes can be selected by the transition possibility:
$$
P _ { u , v } = \left\{ \begin{array} { l l } \frac{1}{d_u} , & if  \  (u, v )\in E, \\ 0,& otherwise, \end{array} \right.
$$
where $P _ { u , v }$ is the transition possibility from node $u$ to $v$, and $d_u$ is the degree of node $u$.

We retreat to \emph{reject sampling}-like methods to guarantee the second and third properties.
Here we introduce the Kullback–Leibler divergence (KL divergence) as a metric to measure the difference between two different distributions. Aiming to get 100 subgraphs with relatively low KL divergence of node labels' distribution (``Node KL'') and edge categories' distribution (``Edge KL''), we set a pre-defined threshold to decide whether to accept a sampled subgraph or not. 
The comparison of the results before and after adding the threshold is shown in Table \ref{tab:RW+ statistics}.

\begin{table}[h]
    \centering
    \caption{KL divergence of the sampling results.}
    \label{tab:RW+ statistics}
    \scalebox{0.65}{
    \begin{tabular}{lcccc}
    \toprule 
        & AMiner& Facebook & NELL & Flickr \\  
    \midrule
         
     Node KL & 0.0186 $\pm$ 0.0107 & 0.1306 $\pm$ 0.0427 & 0.4393 $\pm$ 0.1008 & 0.0060 $\pm$ 0.0025\\
     Edge KL & 0.0189 $\pm$ 0.0149 & 0.0284 $\pm$ 0.0121 & 0.2796 $\pm$ 0.0678 & 0.0046 $\pm$ 0.0015\\
    \midrule
    \emph{+Thresholds}\\
    \midrule
     Node KL  & 0.0123 $\pm$ 0.0024 & 0.0326 $\pm$ 0.0064 & 0.3184 $\pm$ 0.0243 & 0.0041 $\pm$ 0.0006\\
     Edge KL & 0.0062 $\pm$ 0.0008 & 0.0243 $\pm$ 0.0120 & 0.2068 $\pm$ 0.0179 & 0.0021 $\pm$ 0.0003\\

    \bottomrule
    \end{tabular}
    }
\end{table}

\begin{table}[t]

\centering
 \caption{Evaluation Results of GNNs on IGB.}
 \label{tab:result}
 \small
\scalebox{0.9}{
\begin{tabular}{lccccc}
    \toprule & AMiner & Facebook & NELL & Flickr & Avg\\
    \midrule
    Grand  & 82.5$\pm$0.8 & 88.9$\pm$1.0 & 84.4$\pm$1.1 & 44.3$\pm$0.8  & 75.0 \\
    GCN & 76.5$\pm$1.1 & 87.9$\pm$1.0 & 93.9$\pm$0.7 & 41.9$\pm$1.3 & 75.1 \\
    GAT & 78.8$\pm$1.0 & 88.3$\pm$1.2 & 91.1$\pm$1.2 & 43.1$\pm$1.3 & 75.3 \\
    GraphSAGE & 81.6$\pm$0.8 & 87.2$\pm$1.1 & \textbf{94.9}$\pm$\textbf{0.6} & 43.4$\pm$0.9 & 76.8 \\
    APPNP & 87.0$\pm$1.0 & 88.0$\pm$1.3 & 93.0$\pm$0.8 & 44.6$\pm$0.9 & 78.2 \\
    MixHop & 86.1$\pm$1.1 & 89.1$\pm$0.9 & 94.7$\pm$0.7 & 43.5$\pm$1.2 & 78.4 \\
    GCNII & \textbf{88.4}$\pm$\textbf{0.6} & \textbf{89.5}$\pm$\textbf{0.9} & 91.5$\pm$1.0 & \textbf{44.7}$\pm$\textbf{0.8} & \textbf{78.5} \\ 

    \bottomrule
  \end{tabular}
   }
\vspace{-2mm}%
\end{table}




\subsection{Benchmarking Results}
We evaluate 7 representative GNNs on IGB: Grand~\cite{Grand}, GCNII~\cite{chen2020simple}, APPNP~\cite{klicpera2019predict}, GAT~\cite{Velickovic2018GraphAN}, GCN~\cite{kipf2016semi-GCN}, GraphSAGE~\cite{Hamilton2017InductiveRL} and MixHop~\cite{MixHop}. The results are shown in Table~\ref{tab:result}.

To evaluate the GNN model, we first define a search scope for each hyper-parameter. The scope is carefully chosen in order to include the best values on all the datasets. For each model, the best hyper-parameters in its original paper and the best hyper-parameters reported by CogDL~\cite{Cen2021CogDLAE} are usually included in the search scopes. After that, we use our IGB benchmark to evaluate each model under the setting introduced in the section~\ref{sec:pipe}.

\subsection{The Stability of IGB}
\begin{figure}[h]
    \centering
    \includegraphics[width=\linewidth]{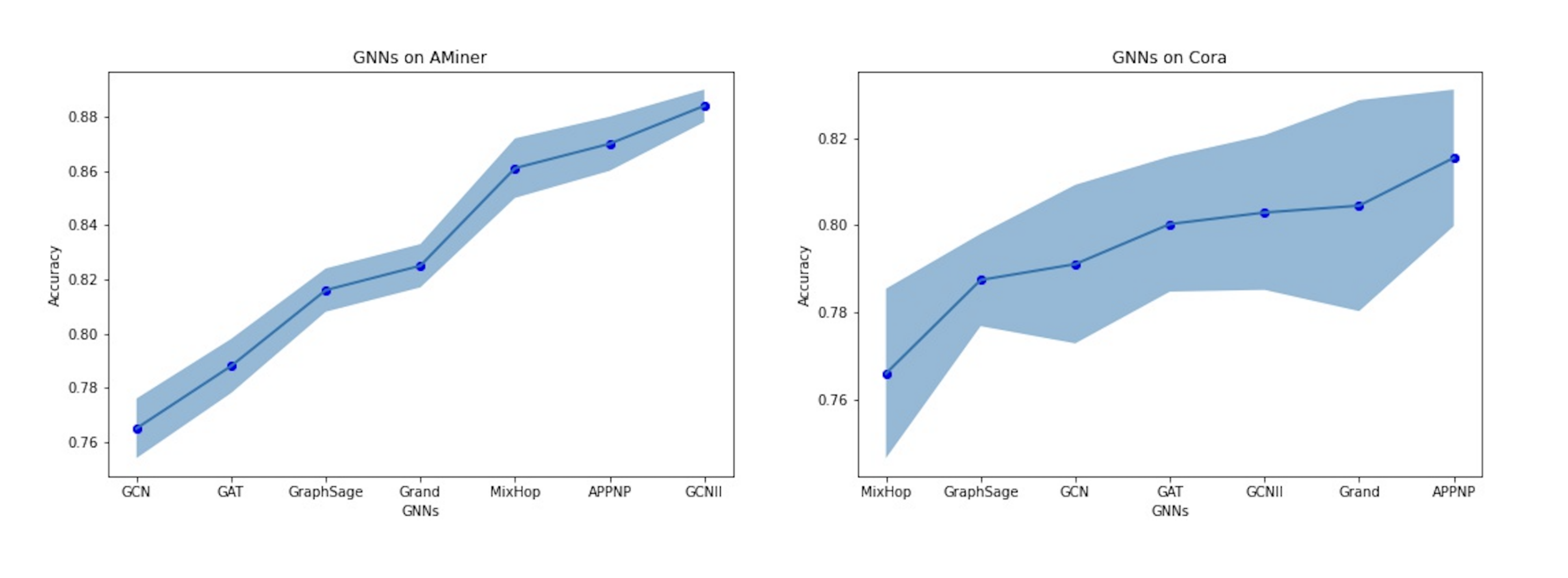}
    \caption{Accuracy of GNNs on AMiner and Cora. The blue area is the fluctuation range of the test accuracy. The results are based on 100 runs on AMiner's subgraphs or Cora's random splits. }
    \label{fig:gnns_on_aminer}
\end{figure}

We verify the stability of IGB in two ways. Firstly, we verify its stability when evaluating models on different graphs, as each IGB dataset contains 100 nearly i.i.d. graphs. Specifically, we compare the variances of the accuracies on 100 AMiner's subgraphs (IGB style) and on 100 Cora's random data splits (Cora style). The result shown in Figure~\ref{fig:gnns_on_aminer} strongly suggests that the evaluation on IGB is more stable than Cora style, even though each AMiner graph uses random data split.

Secondly, we focus on IGB's stability when evaluating models with different random seeds. In a stable benchmark, the ranks of different models should not easily change when changing the random seed. To verify this, we use the ``inversion number'' \footnote{The definition of ``Inversion'' can be seen in \href{https://en.wikipedia.org/wiki/Inversion_(discrete_mathematics)}{Wikipedia}.} of the ranking as a metric. Specifically, we evaluate the seven models using ten different random seeds, providing ten ranking sequences $S_i (i\in{[1,10]})$. On each ranking sequence, we sort models according to their accuracy. The ranking sequence of the first seed is used as a \emph{reference} sequence $S_1 = \{m_1, m_2...m_7\}$, where $m_i$ is a GNN model. We call $m_j>m_k$ if and only if $m_j$ ranks higher than $m_j$ on $S_1$. For another sequence $S_i, i\neq 1$, if $m_j>m_k$, but $m_j$ ranks lower than $m_k$ on $S_i$, we call $(j,k)$ an inversion pair in $S_i$. ``Inversion number'' is the number of inversion pairs in all sequences $S_i, i\neq 1$. Therefore, a high ``inversion number'' indicates a high instability of evaluation with different seeds.
The result is reported in Table \ref{reverse_order_pair}. IGB has a significantly smaller inversion number than Cora, CiteSeer, and PubMed, demonstrating its strong stability.

\begin{table}[h]
\centering
\caption{The inversion numbers of the ranking sequences using 10 different random seeds. Smaller inversion number indicates better stability.}
\label{reverse_order_pair}
\scalebox{0.8}{
\begin{tabular}{ccccccc}
    \toprule
     Cora & CiteSeer & PubMed & AMiner & Facebook & NELL & Flickr \\
     \midrule
     67 & 45 & 107 & 0 & 9 & 0 & 5 \\
    \bottomrule
\end{tabular}
}
\vspace{-3mm}%
\end{table}

\section{Discussion}
\textbf{Is limiting the number of hyper-parameters a good way to solve the over-tuning problems? } In section 2, we illustrate the power of over-tuning, where the improvements basically correlate with the number of hyper-parameters. However, if we set a hard limit for the number of hyper-parameters, complicated optimizers with many hyper-parameters, e.g. Adam~\cite{kingma2014adam}, will not be encouraged due to this limit. The models will also be encouraged to investigate more influential hyper-parameters to utilize the labels in the validation set under the limited budget of hyper-parameters. Therefore, the most fundamental solution is to change the evaluation setting as IGB.\vspace{2mm}\\

\noindent\textbf{What is the best GNN? }In the results of IGB, GCNII performs the best. However, the performance differs in different datasets. For example, the sota method on Citeseer, GRAND, performs badly on NELL and thus gets a low average score, because NELL is a knowledge graph, whose distribution is quite different from that of citation networks. Will a GNN be good at all kinds of graphs, or do we need to design different GNNs for different categories of graphs? 
\section{Conclusion} 
In this paper, we revisit the setting of semi-supervised learning on graphs, identify the over-tuning problem, and prove its significance via the experiments of ValidUtil. To solve it, we propose a new benchmark, IGB, with a more reasonable evaluation pipeline. To further increase evaluation stability, we propose a sampling algorithm based on RW. GNNs are evaluated on the new benchmark, and the results demonstrate a stable performance rank. We expect that IGB can benefit the graph learning community by stabilizing the evolving path of GNNs in the future.

\clearpage

\bibliographystyle{named}
\bibliography{ijcai22}

\end{document}